\documentclass{article}
\PassOptionsToPackage{numbers,sort&compress}{natbib}

\usepackage[final]{mlcd_2021}

\usepackage[utf8]{inputenc} 
\usepackage[T1]{fontenc}    
\usepackage{hyperref}       
\usepackage{url}            
\usepackage{booktabs}       
\usepackage{amsfonts}       
\usepackage{nicefrac}       
\usepackage{microtype}      
\usepackage{xcolor}         
\usepackage{tabularx}
\usepackage{tabulary}
\usepackage{multirow}
\newcolumntype{Y}{>{\centering\arraybackslash}X}

\usepackage{physics}
\usepackage{amsmath}
\usepackage{tikz}
\usepackage{mathdots}
\usepackage{yhmath}
\usepackage{cancel}
\usepackage{color}
\usepackage{siunitx}
\usepackage{array}
\usepackage{multirow}
\usepackage{amssymb}
\usepackage{tabularx}
\usepackage{extarrows}
\usepackage{booktabs}
\usetikzlibrary{fadings}
\usetikzlibrary{patterns}
\usetikzlibrary{shadows.blur}
\usetikzlibrary{shapes}

\usepackage{cleveref}
\usepackage{bm}

\usepackage{array}
\newcolumntype{C}[1]{>{\centering\arraybackslash}p{#1}}

\title{Physically Embodied Deep Image Optimisation}

\author{%
  Daniela Mihai\thanks{Authors contributed equally}\hspace{0.4em}\thanks{Vision, Learning and Control Group, Electronics and Computer Science, University of Southampton, \{adm1g15, jsh2\}@ecs.soton.ac.uk} \\
   \And
   Jonathon Hare\footnotemark[1]\hspace{0.4em}\footnotemark[2]\\
}

\begin{document}

\maketitle


\begin{figure}[h!]
    \centering
    
    \includegraphics[width=\textwidth]{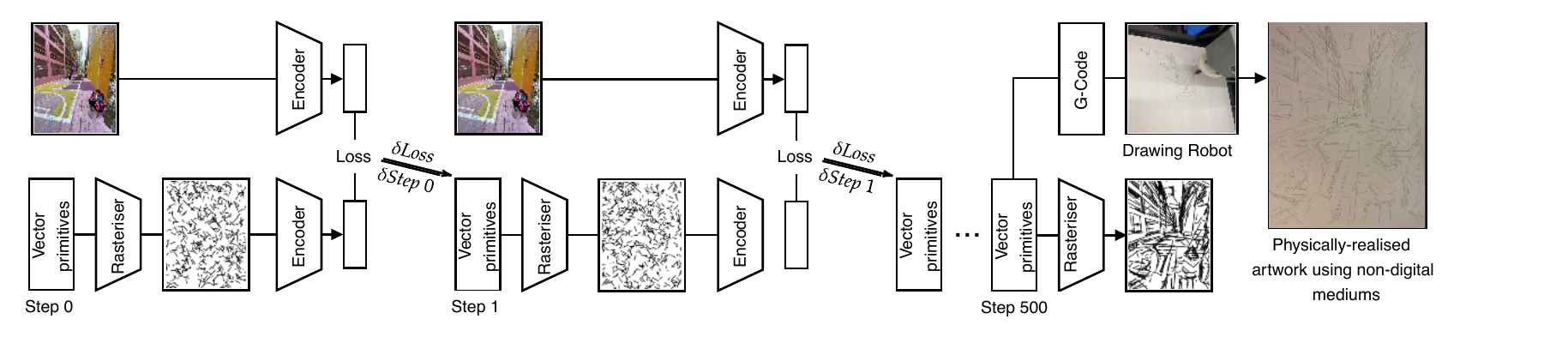}
    \caption{We create physical sketches by learning programs to control a drawing robot. A differentiable rasteriser is used to optimise sets of drawing strokes to match an input image, using deep networks to provide an encoding for which we can compute a loss. The optimised drawing primitives can then be translated into G-code commands which command a robot to draw the image using drawing instruments such as pens and pencils on a physical support medium.}
    \label{fig:overview}
\end{figure}

\section{Introduction}
Recent progress in differentiable rasterisation of primitives (points, lines and curves) on a pixel raster allow us to build machines that turn digital raster images into continuous parametric representations and back into rasterised images again~\citep{mihai2021differentiable}. Using such an approach, we can model the physical act of drawing with a pen on paper. 

This approach differs from other methods which use neural networks to generate painting-like images, such as style transfer \citep{gatys2015neural} or GANs\citep{elgammal2017can}, which essentially map textures across pixel grids and do not model the physical process of drawing. Stroke based-rendering with some optimisation goal has been studied before the deep learning era \citep{hertzmann2003survey}. In the field of robot art, Pix18~\citep{lipsonPix18} excels at creating oil paintings with its own art subject, and with minimal human intervention.
While recent advances to image generation have focused on reinforcement learning \citep{ganin2018synthesizing} or neural renderers \citep{zheng2018strokenet,nakano2019neuralpainters}
, the optimisation of brushstrokes against loss functions parameterised by deep networks, which can then be directly interpreted by a drawing robot has been missing from the field.

\section{Optimisation of Stroke Primitive Parameters with Different Losses}\label{sec:optimisationmethod}
With the machinery defined in \citet{mihai2021differentiable} it is possible to define a complete system that takes the parameters describing stroke primitives and rasterises those strokes into an image. As shown in \cref{fig:overview}, by introducing a (pretrained) neural network to extract features from both the complete rasterised image and a fixed target image, coupled with an appropriate loss function computed between those features, it becomes possible to compute gradients with respect to the parameters of the primitives that created the rasterised image. Minimising this loss will adapt the underlying primitives to ``shapes'' that maximise the feature similarity of the target image. If the rasterisation function is differentiable with respect to the stroke primitives, then the minimisation problem can be solved using gradient descent. 
Further, the optimised stroke parameters can be directly transformed into instructions that control a robot (see \cref{app:robot}) that can manipulate a drawing instrument like a pen or pencil over a support medium. This allows us to produce physical artefacts directly from the  model.

In our experiments, we optimise the underlying primitive parameters by minimising either the MSE loss directly against the image pixels (the encoder network is the identity function) or a loss built upon features extracted from the intermediate layers of a deep convolutional network. For the latter loss which is inherently differentiable, features are extracted using a VGG16 network pretrained on StylisedImageNet dataset (SIN) \citep{geirhos2018imagenetTexture} which has a shape bias, or the regular ImageNet dataset~\citep{deng2009imagenet} which has been shown to be texture-biased.

\section{Case Studies}
We provide two case studies to demonstrate the effectiveness of our approach. Firstly we demonstrate the effects can be achieved by transforming photographs to line drawings by optimising against different intermediate representations of a deep network. Secondly, we demonstrate a variation of the idea of neural style transfer \citep{gatys2015neural} where instead of generating raster images we directly create the underlying stroke information that allows an image to be drawn under different physical drawing constraints.

\paragraph{How does my deep network think an image should be sketched?}

We can use our technique to explore how different internal representations of pretrained encoder networks manifest themselves in terms of illustrations of photographs as shown in \cref{fig:perceptions}. Interestingly this allows us to explore different artistic effects that arise, but also to qualitatively look at the variations between differing inductive biases and loss formulations.

\begin{center}
    \captionsetup{type=figure}
    \centering
    \includegraphics[height=1.5cm]{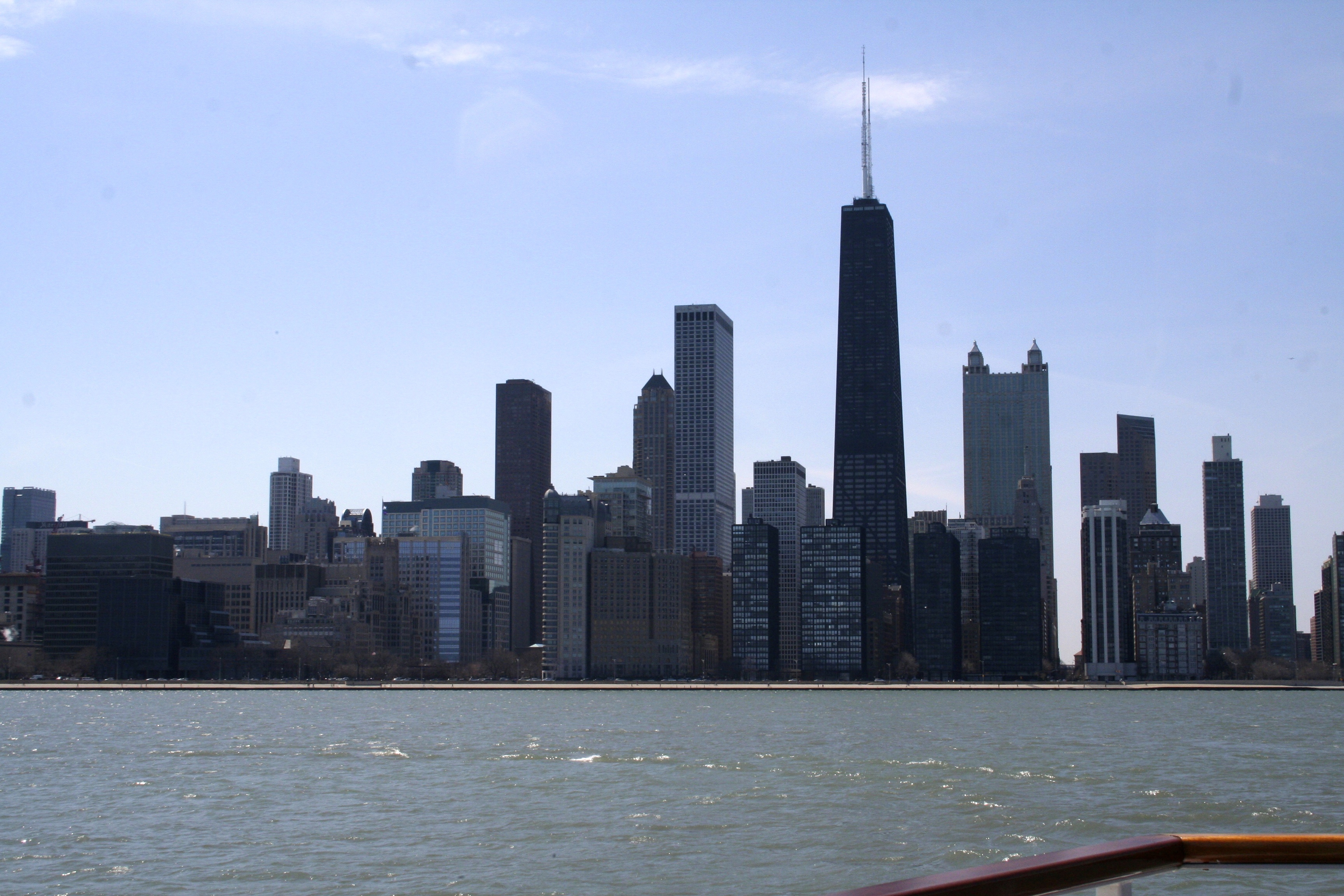}
    \includegraphics[height=1.5cm]{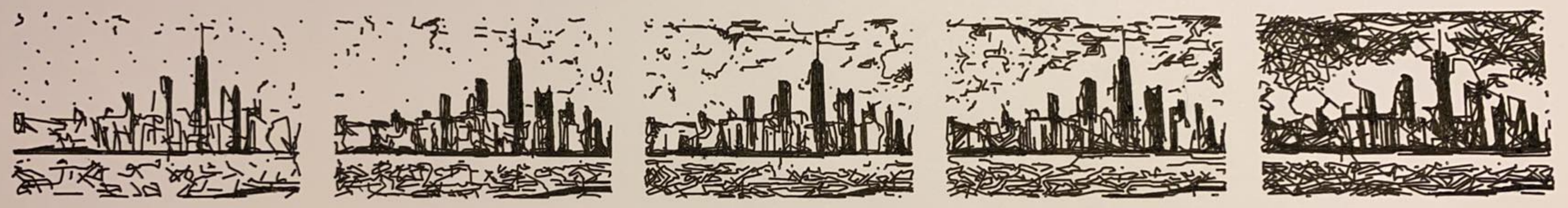}
    
  \caption{Illustrations created from VGG-16 networks with SIN weights at a range of different depths (early layers left, later layers right). Each image was created using 1000 straight lines.}\label{fig:perceptions}
\end{center}

\paragraph{What if Edvard Munch's `The Scream' was painted in the Pointillism technique? Or Cubism?}\Cref{fig:munchfigures} showcases the results of optimising different primitives to fit a photo of the original ``The Scream" by Edvard Munch \footnote{Photo taken from Wikipedia \url{https://en.wikipedia.org/wiki/The_Scream}}. We display results by optimising 2000 uniformly coloured points (\cref{fig:munch1000pointsmse,fig:munchpointssin}) and 1000 coloured lines (\cref{fig:munchlinessin,fig:munch1000linesmse}) to fit the original image by minimising either an MSE or a perceptual loss (with SIN-pretrained VGG16). The contrast between the images with the same type of primitive parametrisation, but using a different loss, is striking. The perceptual loss captures the shape information rather well, while moving away from the colour or texture scheme of the original or the variant realised with MSE loss. The copies thus produced could be tagged as belonging to the Fauvism art movement. Changing the parametrisation of the drawings in the method described in \cref{sec:optimisationmethod} gives us an idea of what the painting would've looked like if it were to have been drawn in a Pointillist style (\cref{fig:munch1000pointsmse,fig:munchpointssin}), or a more abstract, Cubism-like style (\cref{fig:munchlinessin,fig:munch1000linesmse}), while at the same time, making it possible for the drawing to be physically produced on paper by a drawing agent. More examples can be found in \cref{app:morecasestudies}.
\begin{center}
    \captionsetup{type=figure}
    \centering
     \begin{subfigure}[t]{.19\columnwidth}
        \centering
        \includegraphics[width=0.75\columnwidth]{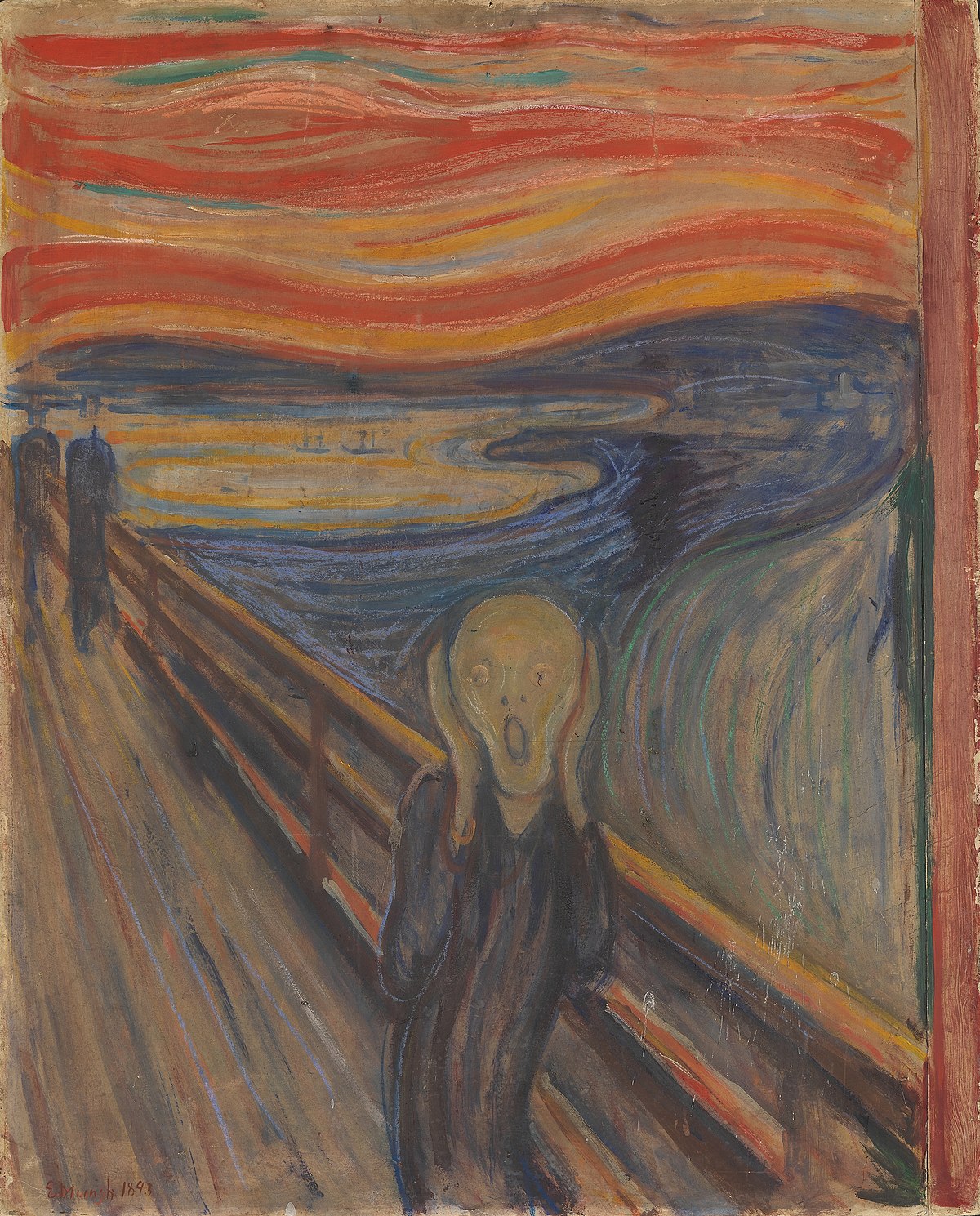}
        \caption{'The Scream'}\label{fig:originalthescream}
    \end{subfigure}
    \begin{subfigure}[t]{.19\columnwidth}
        \centering
        \includegraphics[width=0.75\columnwidth]{images/creativity-workshop/munch-thescream-sop-sinweights-2000points.pdf}
    \caption{2000 Points}\label{fig:munchpointssin}
    \end{subfigure}
    \begin{subfigure}[t]{.19\columnwidth}
        \centering
        \includegraphics[width=0.75\columnwidth]{images/creativity-workshop/munch-thescream-sop-sinweights-1000lines.pdf}
        \caption{1000 Lines}\label{fig:munchlinessin}
  \end{subfigure}
    \begin{subfigure}[t]{.19\columnwidth}
        \centering
        \includegraphics[width=0.75\columnwidth]{images/creativity-workshop/munch-thescream-mse-2000points.pdf}
    \caption{2000 Points}\label{fig:munch1000pointsmse}
    \end{subfigure}
    \begin{subfigure}[t]{.19\columnwidth}
        \centering
        \includegraphics[width=0.75\columnwidth]{images/creativity-workshop/munch-thescream-mse-1000lines.pdf}
        \caption{1000 lines}\label{fig:munch1000linesmse}
  \end{subfigure}
  \caption{Munch's `The Scream'(\cref{fig:originalthescream}) reduced to points and straight lines by gradient decent using LPIPS loss with SIN-pretrained VGG16 (\cref{fig:munchpointssin,fig:munchlinessin}) and MSE loss  (\cref{fig:munch1000pointsmse,fig:munch1000linesmse}). }\label{fig:munchfigures}
\end{center}


\bibliographystyle{plainnat}
\bibliography{main}

\appendix
\section{Our drawing robot}\label{app:robot}

\begin{figure}[h]
    \centering
    \begin{subfigure}[t]{0.49\textwidth}
    \includegraphics[width=\textwidth]{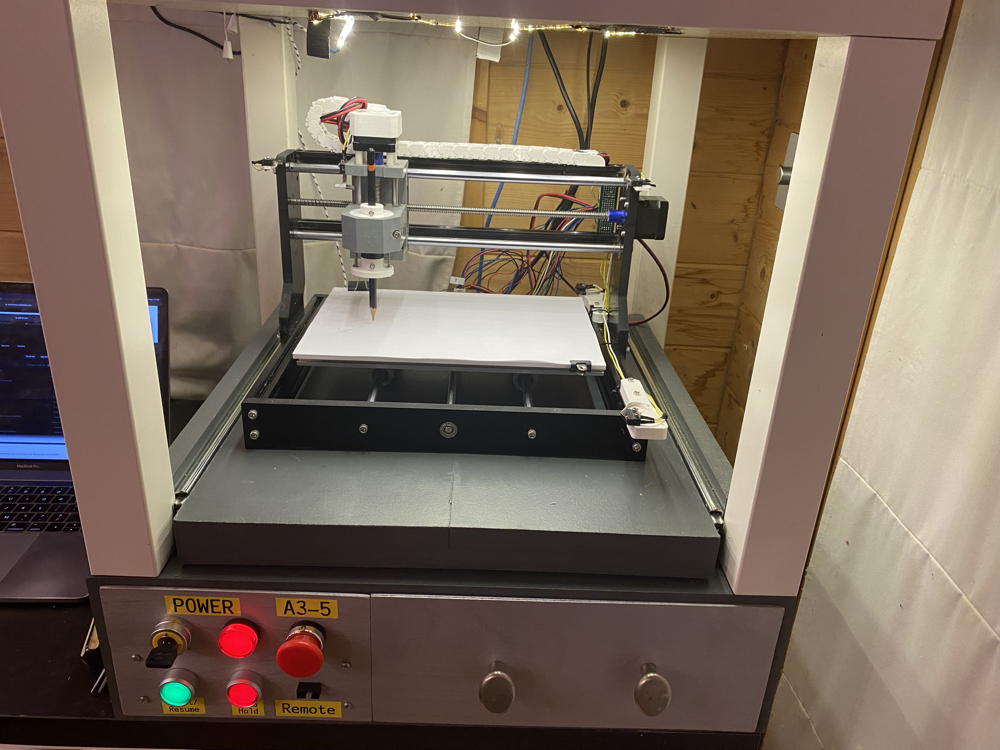}    
    \end{subfigure}\hfill%
    \begin{subfigure}[t]{0.49\textwidth}
    \includegraphics[width=\textwidth]{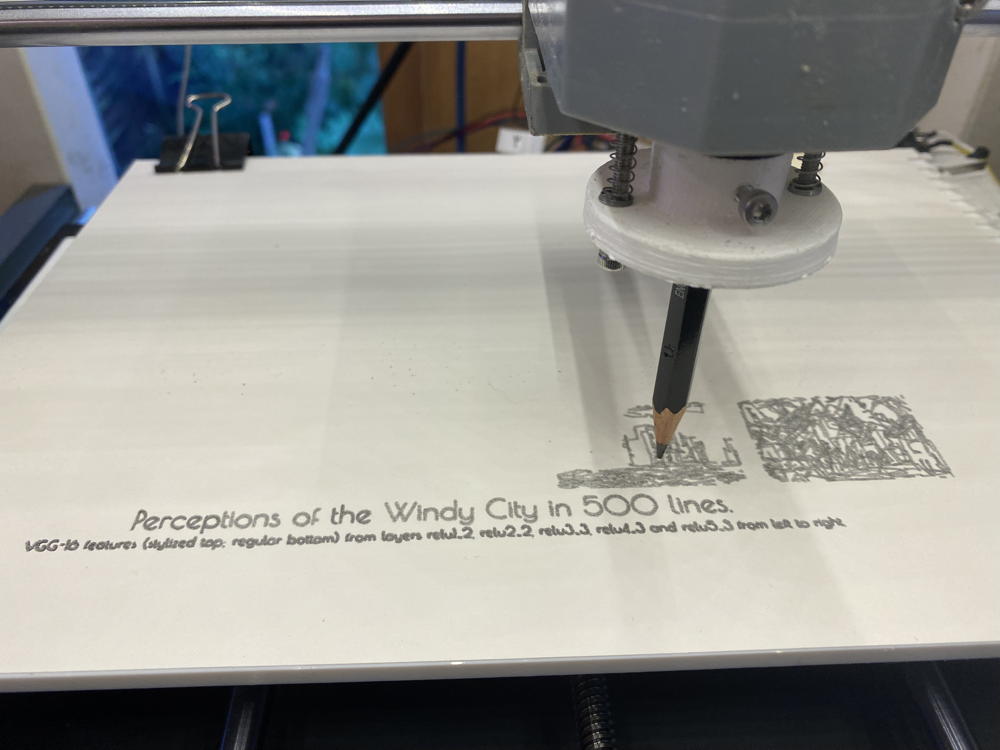}    
    \end{subfigure}
    
    \caption{Photos of our drawing robot in action.}
    \label{fig:robot}
\end{figure}

For producing our artwork we used a custom modified gantry-style robot to manipulate a pen or pencil over a paper support medium as illustrated in \cref{fig:robot}. We used \verb|vpype| (\url{https://github.com/abey79/vpype}) to optimise the stroke order produced by our network to minimise unnecessary movements between strokes, and \verb|juicy-gcode| (\url{https://hackage.haskell.org/package/juicy-gcode}) to convert the strokes into gcode instructions that could be performed by the drawing robot.

\section{More Examples of Image Optimisation }\label{app:morecasestudies}

\begin{figure*}[ht]
    \centering
    \begin{subfigure}[t]{.19\columnwidth}
        \centering
        \includegraphics[width=0.8\columnwidth]{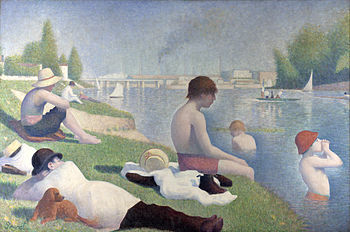}
    \caption{Original}\label{fig:baigneursorig}
    \end{subfigure}
    \begin{subfigure}[t]{.19\columnwidth}
        \centering
        \includegraphics[width=0.8\columnwidth]{images/creativity-workshop/seurat-baigneurs-mse-1000lines.pdf}
    \caption{1000 Lines}\label{fig:baigneurs1000lpips}
    \end{subfigure}
    \begin{subfigure}[t]{.19\columnwidth}
        \centering
        \includegraphics[width=0.8\columnwidth]{images/creativity-workshop/seurat-baigneurs-mse-500crs.pdf}
        \caption{500 CRS}\label{fig:baigneurs500lpips}
    \end{subfigure}
    \begin{subfigure}[t]{.19\columnwidth}
        \centering
        \includegraphics[width=0.8\columnwidth]{images/creativity-workshop/seurat-baigneurs-sop-sinweights-1000lines.pdf}
    \caption{1000 Lines}\label{fig:seurat1000lines}
    \end{subfigure}
    \begin{subfigure}[t]{.19\columnwidth}
        \centering
        \includegraphics[width=0.8\columnwidth]{images/creativity-workshop/seurat-baigneurs-sop-sinweights-500crs.pdf}
        \caption{500 Curves}\label{fig:seurat500curves}
  \end{subfigure}
  \begin{subfigure}[t]{.19\columnwidth}
        \centering
        \includegraphics[width=0.8\columnwidth]{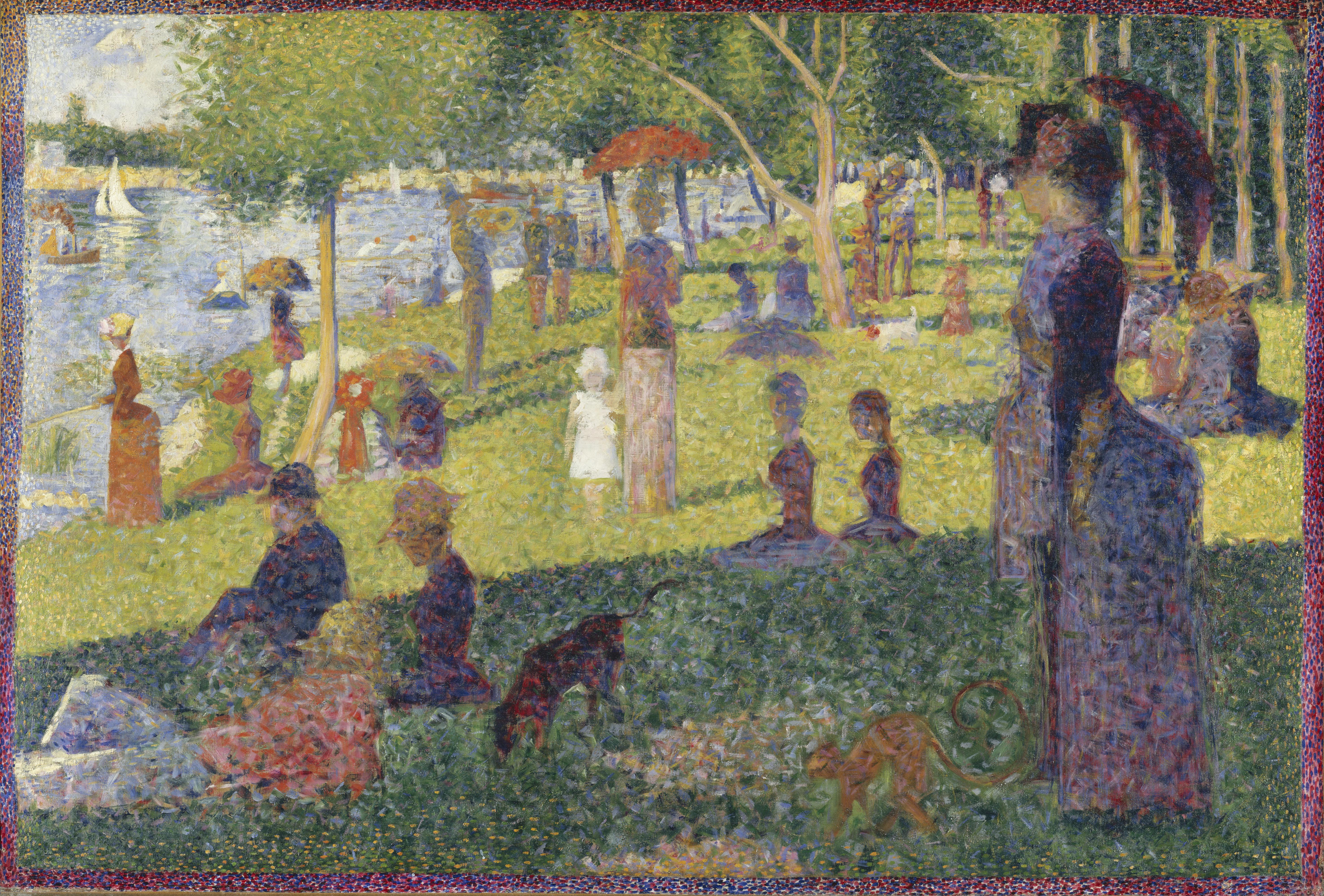}
    \caption{Original}\label{fig:parkoriginal}
    \end{subfigure}
    \begin{subfigure}[t]{.19\columnwidth}
        \centering
        \includegraphics[width=0.8\columnwidth]{images/creativity-workshop/seurat-park-mse-1000lines.pdf}
    \caption{1000 Lines}\label{fig:park1000lpips}
    \end{subfigure}
    \begin{subfigure}[t]{.19\columnwidth}
        \centering
        \includegraphics[width=0.8\columnwidth]{images/creativity-workshop/seurat-park-mse-500crs.pdf}
        \caption{500 CRS}\label{fig:park500lpips}
    \end{subfigure}
    \begin{subfigure}[t]{.19\columnwidth}
        \centering
        \includegraphics[width=0.8\columnwidth]{images/creativity-workshop/seurat-park-sop-sinweights-1000lines.pdf}
    \caption{1000 Lines}\label{fig:parksop1000lines}
    \end{subfigure}
    \begin{subfigure}[t]{.19\columnwidth}
        \centering
        \includegraphics[width=0.8\columnwidth]{images/creativity-workshop/seurat-park-sop-sinweights-500crs.pdf}
        \caption{500 Curves}\label{fig:parksop500curves}
  \end{subfigure}
  \caption{Photographs of Seurat's `Une Baignade, Asni`eres' (taken from \url{https://en.wikipedia.org/wiki/Bathers_at_Asnieres}) and `A Sunday Afternoon on the Island of La Grande Jatte' (\url{https://en.wikipedia.org/wiki/A_Sunday_Afternoon_on_the_Island_of_La_Grande_Jatte}) reduced to straight lines or Catmull-Rom splines by optimising with an MSE loss (\cref{fig:baigneurs1000lpips,fig:baigneurs500lpips,fig:park1000lpips,fig:park500lpips}) or a perceptual loss using features extracted by a SIN-pretrained VGG16 (\cref{fig:seurat1000lines,fig:seurat500curves,fig:parksop1000lines,fig:parksop500curves}).} \label{fig:seurat}
\end{figure*}

In this section, we provide more examples of our approach for image-based optimisation.
\Cref{fig:seurat} shows results of optimising 1000 lines and 500 curves parameterised as Catmull-Rom splines against two of Seurat's paintings, `Une Baignade at Asni`eres' and `A Sunday Afternoon on the Island of La Grande Jatte'. This also shows the differences in optimising with a perceptual loss based on features extracted with a VGG network pretrained on SIN against a simple reconstruction loss.

\begin{figure*}[ht]
    \centering
    \begin{subfigure}[t]{.24\columnwidth}
        \centering
        \includegraphics[width=0.8\columnwidth]{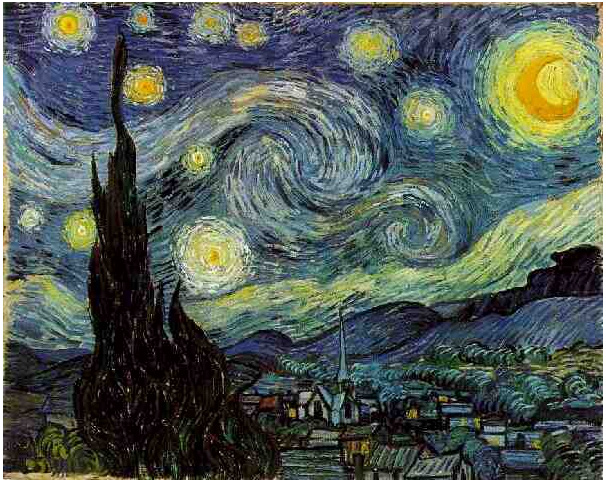}
    \caption{Original}\label{fig:starrynightorig}
    \end{subfigure}
    \begin{subfigure}[t]{.24\columnwidth}
        \centering
        \includegraphics[width=0.8\columnwidth]{images/creativity-workshop/vangogh-starrynight-sop-sinweights-500crs.pdf}
    \caption{500 Curves}\label{fig:starrynight500curves}
    \end{subfigure}
    \begin{subfigure}[t]{.24\columnwidth}
        \centering
        \includegraphics[width=0.8\columnwidth]{images/creativity-workshop/vangogh-starrynight-sop-sinweights-1000lines.pdf}
    \caption{1000 Lines}\label{fig:starrynight1000lines}
    \end{subfigure}
    \begin{subfigure}[t]{.24\columnwidth}
        \centering
        \includegraphics[width=0.8\columnwidth]{images/creativity-workshop/vangogh-starrynight-sop-sinweights-2000points.pdf}
        \caption{2000 Points}\label{fig:starrynight2000points}
    \end{subfigure}
  \caption{Van Gogh's `The Starry Night' (taken from \url{https://en.wikipedia.org/wiki/The_Starry_Night}) reduced to curves, lines and points by optimisation using a perceptual loss with SIN-pretrained VGG16.}\label{fig:vangogh}
\end{figure*}

Similarly, \cref{fig:vangogh} shows a comparison between the image optimisation results using different primitive parametrisations by minimising a perceptual loss.

\end{document}